# Multilingual Non-Factoid Question Answering with Answer Paragraph Selection

**Ritwik Mishra[1]**, **Sreeram Vennam[2]**, **Rajiv Ratn Shah[1]** and **Ponnurangam Kumaraguru[2]**

[1] Indraprastha Institute of Information Technology, Delhi
`{ritwikm, rajivratn}@iiitd.ac.in`
[2] International Institute of Information Technology, Hyderabad
`sreeram.vennam@students.iiit.ac.in, pk.guru@iiit.ac.in`

## Abstract

Most existing Question Answering Datasets (QuADs) primarily focus on factoid-based short-context Question Answering (QA) in high-resource languages. However, the scope of such datasets for low-resource languages remains limited, with only a few works centered on factoid-based QuADs and none on non-factoid QuADs. Therefore, this work presents MuNfQuAD, a multilingual QuAD with *non-factoid* questions. It utilizes interrogative sub-headings from BBC news articles as questions and the corresponding paragraphs as silver answers. The dataset comprises over 578K QA pairs across 38 languages, encompassing several low-resource languages, and stands as the largest multilingual QA dataset to date. Based on the manual annotations of 790 QA-pairs from MuNfQuAD (golden set), we observe that 98% of questions can be answered using their corresponding silver answer. Our fine-tuned Answer Paragraph Selection (APS) model outperforms the baselines. The APS model attained an accuracy of 80% and 72%, as well as a macro F1 of 72% and 66%, on the MuNfQuAD testset and the golden set, respectively. Furthermore, the APS model effectively generalizes a certain language within the golden set, even after being fine-tuned on silver labels. We also observe that the fine-tuned APS model is beneficial for reducing the context of a question. These findings suggest that this resource would be a valuable contribution to the QA research community.

## 1 Introduction

A typical Question Answering Dataset (QuAD) conventionally comprises question-answer pairs (Baudiš and Šedivỳ, 2015; Berant et al., 2013). However, certain QuADs are characterized by an additional component called *evidence* or *context* accompanying each question. This contextual information is expected to provide sufficient details to address the corresponding question, leading to these QuADs being referred to as Reading-Comprehension (RC) datasets as well. The majority of RC datasets focus on factoid answers, typically short phrases or named entities (Soleimani et al., 2021). For example, consider a factoid question, *Who was the first Prime Minister of India?*, with the corresponding factoid answer, *Jawaharlal Nehru*.

As compared to factoid questions, non-factoid questions have long descriptive answers consisting of multiple sentences or paragraphs. Extending the earlier example, a non-factoid question could be framed as *How did Jawaharlal Nehru become the first Prime Minister of India?* Evidence suggests that modern search engines are unable to answer non-factoid questions effectively (Cambazoglu et al., 2021a). Moreover, even humans find it difficult to answer non-factoid questions (Bolotova et al., 2022). In order to automatically answer non-factoid questions, large non-factoid QuADs are needed to fine-tune Question-Answering (QA) models. Multilingual QA models face additional challenges due to the lack of such resources, motivating the development of a multilingual QA dataset specifically designed for non-factoid questions.

In this study, we automatically extract Question-Answer pairs and their corresponding news articles from the British Broadcasting Corporation (BBC) website in multiple languages[1]. Except for the golden set, the dataset is not manually annotated since it relies on the hypothesis put forth by Soleimani et al. (2021) that *all the paragraphs succeeding an interrogative subheading contains its answer*. Therefore, we refer to this dataset as having silver labels/answers. Previous studies have indicated that silver labels have proven beneficial for constructing text classifiers in domains with limited availability of gold labels, such as legal (Neer-bek et al., 2020), medical (Nowak et al., 2023), and news (Cripwell et al., 2023) domains. An evaluation contrasting the silver labels against the gold labels reveals that 98% of the questions were effectively answered by their silver answers. Our main contributions are outlined as follows:

1. We release MuNfQuAD[2], a multilingual question-answering dataset partitioned into train, validation, and test splits. It contains more than 578K Question-Answer pairs in 38 different languages.

2. We also release multilingual Answer Paragraph Selection models fine-tuned on MuNfQuAD, leveraging base variants of different pretrained encoders.

---
[1] `https://www.bbc.co.uk/ws/languages`

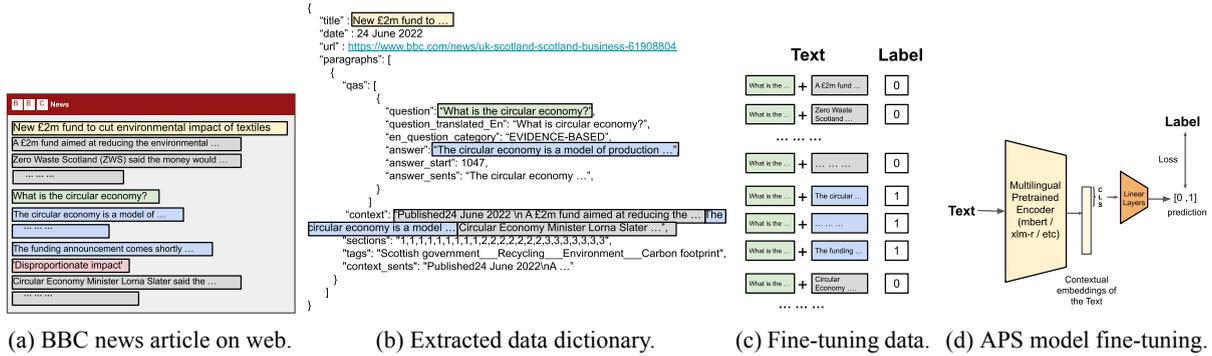

(a) BBC news article on web.  (b) Extracted data dictionary.  (c) Fine-tuning data.  (d) APS model fine-tuning.

Figure 1: An illustration depicting the data collection process and fine-tuning of the Answer Paragraph Selection (APS) model. (a) Presents a BBC news article accessed via the URL: https://www.bbc.com/news/uk-scotland-scotland-business-61908804. We have marked interrogative subheadings with green boxes, non-interrogative subheadings with red boxes, and silver answer paragraphs with blue boxes only for illustrative purposes. The scraped data dictionary is shown in (b). In subfigure (c), Label-0 indicates that the given paragraph is not included in the silver answer, while Label-1 signifies that it is. APS model architecture and the fine-tuning process are illustrated in (d).

## 2 Related Works

WikiQA (Yang et al., 2015) emerged as an early dataset for automatic QA in English. It extracted questions from Bing query logs and matched them with relevant Wikipedia articles. SQuAD (Rajpurkar et al., 2016) is a benchmark QA dataset in English. Crowdworkers generated questions based on English Wikipedia passages and identified the answer within a short span of text. The most extensive dataset for factoid-based span detection is Natural Questions (Kwiatkowski et al., 2019). It comprises almost 320K questions, each accompanied by a long answer, a short answer, and a complete Wikipedia article as context. Our work closely resembles NLQuAD (Soleimani et al., 2021), but it was designed exclusively for the English language. For a comprehensive review of English QA datasets, readers can refer to Cambazoglu et al. (2021b); Rogers et al. (2023).

The main focus of our work is on multilingual QuADs (mQuADs). An early endeavor in this domain is bAbI (Weston et al., 2016), which contained factoid-based questions and extractive answers in English and Hindi transliterated into Roman script. Gupta et al. (2018) introduced a bilingual Hindi-English QuAD, showing improved QA performance with question classification. Gupta et al. (2019) automatically translated a subset of SQuAD to Hindi, but we observed that a large majority of its answer indices were inaccurate. The XQA (Liu et al., 2019) dataset gathered questions from Wikipedia's "Did you know?" boxes. These questions omitted entity names, which were then employed as factoid answers. The top 10 Wikipedia articles related to the identified entity served as the context for each question. The authors also emphasized the constraints of using translation-based augmentation in QA systems. A subset of the SQuAD dataset was manually translated into ten languages, creating XQuAD (Artetxe et al., 2020).

MLQA (Lewis et al., 2020) engaged crowd workers to generate questions from English Wikipedia articles and provide extractive answers. Subsequently, parallel sentences were extracted from the English article, and the English question-answer pair was manually translated into other languages. TyDi QA (Clark et al., 2020) represents a milestone in multilingual QuADs, focusing on natural questions where question makers are unaware of the answers beforehand. Crowdworkers were encouraged to ask questions out of curiosity, and top-ranked Wikipedia articles were then used for answer paragraphs and minimal answer span labeling. Chaii[3] offers a QuAD with factoid questions and long-context in Tamil and Hindi. Table 1 contains the statistics of different mQuADs. While multilingual datasets like BOLT (Song et al., 2014) and ResPubliQA (Peñas et al., 2010) have associated publications (Chaturvedi et al., 2014; Molino, 2013), we were unable to locate the datasets on the open web.

---

[2]https://github.com/ritwikmishra/MuNfQuAD

| Source | Dataset Name | Avg #tokens | | | Type | #Languages | #Samples ↓ |
| --- | --- | --- | --- | --- | --- | --- | --- |
| | | Q | C | A | | | |
| Google Research India (2021) | Chaii | 7 | 1694 | 2 | Factoid Span Detection | 2 | 1.1K |
| Gupta et al. (2018) | $^\alpha$MMQA | 9 | 314 | 7 | Factoid Span Detection | 2 | 2.7K |
| Artetxe et al. (2020) | XQuAD | 12 | 155 | 4 | Factoid Span Detection | 11 | 13K |
| Gupta et al. (2019) | $^\beta$MQA | 10 | 126 | 2 | Factoid Span Detection | 2 | 36K |
| Asai et al. (2021) | XOR QA | 6 | - | 4 | Open Retreival QA | 7 | 40K |
| Lewis et al. (2020) | MLQA | 8 | 117 | 3 | Factoid Span Detection | 7 | 46K |
| Clark et al. (2020) | TyDiQA-GoldP | 5 | 76 | 4 | Factoid Span Detection | 11 | 54K |
| Clark et al. (2020) | TyDiQA-MinSpan | 5 | 2825 | 4 | Factoid Span Detection | 11 | 78K |
| Clark et al. (2020) | TyDiQA-SelectP | 6 | 2891 | 80 | Answer Paragraph Selection | 11 | 90K |
| Liu et al. (2019) | $^\gamma$XQA | 17 | 5326 | 2 | Factoid Span Detection | 9 | 90K |
| Longpre et al. (2021) | MKQA | 8 | - | 2 | Open Domain QA | 26 | 260K |
| Weston et al. (2016) | $^\delta$bAbI | 5 | 21 | 1 | Factoid Span Detection | 2 | 330K |
| Ours | **MuNfQuAD** | 6 | 909 | 191 | Answer Paragraph Selection | 38 | 578K |

Table 1: Attributes of different multilingual QuADs. For languages like Chinese and Japanese, which lack whitespaces as word boundaries, we used *jieba* and *MeCab* python libraries for tokenization, respectively. Conversely, for other languages, we adopted whitespace-based tokenization. $^\alpha$Question to context mapping was not given in MMQA, so we greedily built a mapping. $^\beta$MQA is a subset of SQuAD automatically translated into Hindi, but ∼92% of its (start, end) indices are incorrect. $^\gamma$We concatenate all (ten) contexts of a question to form a single context in XQA. $^\delta$In the bAbI dataset, we take the text above a question as its context.

## 3 Data Curation

We aimed to create a multilingual QA dataset with non-factoid questions. To achieve this, we utilized automated scraping of the BBC news website, gathering news articles and corresponding question-answer pairs. This study used Python requests and BeautifulSoup libraries to scrape data. For a given language (say Hindi), we ran a scraper on BBC (Hindi) website and another one on the Wayback machine[4] (also called web archive). The seed articles for the BBC website scraper are taken from the latest homepage of BBC (Hindi), whereas Wayback machine scraper starts from the earliest snapshot of the BBC (Hindi) homepage. The scraping approach was designed to extract news articles based on the presence of an interrogative subheading within a webpage. In Figure 1(a), the web interface of a BBC news article is depicted, while Figure 1(b) illustrates the scraped data in dictionary format. However, non-interrogative subheadings present within the article are deliberately omitted from the context. This decision is based on the fact that non-interrogative subheadings typically serve as summaries, convey topic information, or offer descriptive titles (Jang and Kim, 2023). While valuable for contextualizing the content, these aspects may not directly address the specific questions posed in the dataset. Moreover, candidate URLs were sourced from the current webpage by capturing its anchor tags. An interrogative subheading is identified by the trailing question mark (or equivalent symbol of that language) in the subheading text.

We used the multilingual checkpoint of ersatz (Wicks and Post, 2021) to segment the sentences in our dataset. We sorted the questions within MuNfQuAD based on their occurrence frequencies and subsequently translated the most frequent 50 questions from each language into English using Google Translate. It was noted that numerous questions were common among articles with no direct relevance. To resolve this, we made a lexicon (list) of phrases for each language to exclude interrogative subheadings irrelevant to the article. For example, हे वाचलंत का? (*Did you read this?*) is a common interrogative subheading from news articles in Marathi language. Additional examples are provided in Table 9 of Appendix C.

---
[3]https://www.kaggle.com/c/chaii-hindi-and-tamil-question-answering
[4]https://archive.org/

| | |
|---|---|
| Number of Languages | 38 |
| Number of QA pairs | 578K |
| Number of Articles | 294K |
| Number of Unique Questions | 411K |
| Avg. Article Length (Word) | 875 |
| Avg. Paragraph Length (Word) | 17 |
| Avg. Answer Length (Word) | 165 |
| Avg. Question Length (Word) | 6 |
| Avg. Article Length (Sentence) | 51 |
| Avg. Paragraph Length (Sentence) | 1.4 |
| Avg. Answer Length (Sentence) | 9.5 |
| Avg. Question Length (Sentence) | 1.0 |
| Avg. Paragraphs per Article | 35 |
| Avg. Paragraphs per Answer | 6.5 |

Table 2: Overview of MuNfQuAD statistics. We tokenized words using whitespace-based splitting and employed the ersatz library (Wicks and Post, 2021) for multilingual sentence segmentation.

## 3.1 MuNfQuAD Statistics

The presented dataset encompasses over 411K unique question-answer pairs, establishing itself as the most extensive mQuAD. Table 2 provides an overview of diverse statistics related to this dataset. We observed that over 75% of the articles in MuNfQuAD exceed the token limit (512) of traditional multilingual encoders designed for factoid QA models (Kumar et al., 2022; Gaschi et al., 2022). Section 5 discusses the computational constraints faced by state-of-the-art multilingual encoders with higher token limits when processing MuNfQuAD articles. A detailed illustration of word distribution among articles, paragraphs, questions, and answers can be found in Figure 2.

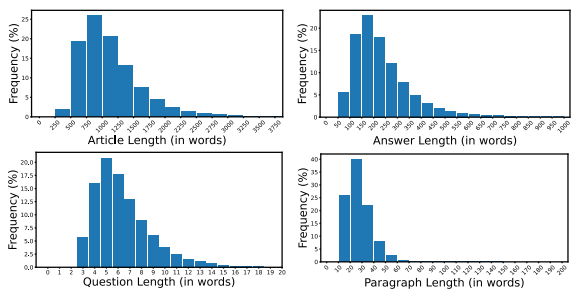

Figure 2: Word frequency distribution in MuNfQuAD.

We conducted web crawling on the BBC news website for all supported languages, resulting in data collection from 38 languages out of the 43 supported. A detailed breakdown of language distribution, along with the corresponding year of the earliest article in MuNfQuAD, is presented in Table 3. Given the highly multilingual nature of the data, we experiment exclusively with QA models that are designed to be multilingual.

To investigate n-gram trends, entity distribution, and question categories, we translated each question within MuNfQuAD to English. This translation was achieved using the *nllb-200-1.3B* model (Team et al., 2022), which boasts the unique capability of translating across 200 languages through a single model. For named entity extraction from English questions, we utilized the spaCy library. Key findings regarding the most frequent n-grams and named entities are presented in Table 4 and Table 5, respectively. To categorize each English question into distinct classes, we utilized the non-factoid question classifier from Bolotova et al. (2022), revealing that more than two-thirds of the questions in the proposed dataset were classified as non-factoid. The similarity in question category distribution between MuNfQuAD and NLQuAD (Soleimani et al., 2021) is illustrated in Figure 4.

## 3.2 MuNfQuAD vs NLQuAD

A key distinction between the two datasets is that MuNfQuAD includes sentence-segmented multilingual data with question categories, which NLQuAD does not. Moreover, we do not assert that BBC articles as question-context qualify as long-context, given that many LLMs capable of processing text longer than BBC articles have since been developed, which were not available when NLQuAD was proposed.

## 4 Answer Paragraph Selection

In the context of a provided question and segmented context paragraphs, the Answer Paragraph Selection (APS) model assigns high confidence scores to paragraphs belonging to the silver answer. The APS model takes as input the concatenation of a question and the $i^{th}$ paragraph ($p_i$) from the context. The output is a probability value ranging from 0 to 1, indicating the likelihood of $p_i$ being an answer to the provided question. The choice of employing an APS model, as opposed to a sliding window Reading-Comprehension model (Soleimani et al., 2021), stems from the APS model's alignment with the Answer Sentence Selection (AS2) approach, which is deemed to be more relevant than the RC approach (Garg et al., 2020; Barlacchi et al., 2022). Xu et al. (2017) demonstrated the effectiveness of the APS component in automatically answering questions whose responses span across non-continuous text segments. As illustrated in Table 1, TyDi QuAD (Clark et al., 2020) similarly in-cludes a separate track for QA models that employ the APS paradigm.

| Region | Africa | | | | | | | | | | | Asia (Central) | |
|---|---|---|---|---|---|---|---|---|---|---|---|---|---|
| Lang. | Oromo | Amharic | French* | Hausa | Igbo | Gahuza | Pidgin** | Somali | Swahili | Tigrinya | Yoruba | Kyrgyz | Uzbek |
| Code | om | am | fr* | ha | ig | rw | en** | so | sw | ti | yo | ky | uz |
| #QA | 5.4k | 2.0k | 4.9k | 2.8k | 1.5k | 3.1k | 5.4k | 5.5k | 6.1k | 3.2k | 1.9k | 2.3k | 2.7k |
| #Articles | 3.3k | 1.0k | 2.1k | 1.5k | 1.1k | 1.8k | 3.4k | 3.1k | 3.3k | 1.9k | 1.4k | 1.3k | 1.6k |
| Start Year | 2017 | 2013 | 2012 | 2013 | 2018 | 2014 | 2017 | 2016 | 2010 | 2017 | 2018 | 2011 | 2010 |
| Region | Asia (Pacific) | | | Asia (South) | | | | | | | | | |
| Lang. | Indonesian | Korean | Vietnamese | Bengali | Gujarati | Hindi | Marathi | Nepali | Pashto | Punjabi | Sinhala | Tamil | Telugu |
| Code | id | ko | vi | bn | gu | hi | mr | ne | ps | pa | si | ta | te |
| #QA | 11k | 4.6k | 10k | 7.9k | 15k | 26k | 21k | 11k | 2.9k | 7.6k | 3.3k | 12k | 15k |
| #Articles | 5.5k | 2.5k | 6k | 4.2k | 8.2k | 14k | 10k | 5.5k | 1.9k | 3.7k | 1.7k | 6.7k | 7.3k |
| Start Year | 2010 | 2017 | 2009 | 2013 | 2017 | 2009 | 2017 | 2014 | 2010 | 2017 | 2012 | 2012 | 2017 |
| Region | Asia (South) | Europe | | | | | | | Latin America | | Middle East | | |
| Lang. | Urdu | Azeri | Russian | Serbian | Turkce | Ukrainian | Cymrufyw | English | Portuguese | Mundo | Arabic | Persian | |
| Code | ur | az | ru | sr | tr | uk | cy | en | pt | es | ar | fa | |
| #QA | 14k | 4.3k | 22k | 21k | 17k | 19k | 8.8k | 14k | 12k | 20k | 11k | 11k | |
| #Articles | 7.4k | 2.3k | 12k | 9.5k | 8.7k | 11k | 2.8k | 6.9k | 5.4k | 11k | 6k | 6.2k | |
| Start Year | 2010 | 2011 | 2010 | 2018 | 2009 | 2009 | 2012 | 2011 | 2011 | 2009 | 2009 | 2008 | |

Table 3: Language distribution in MuNfQuAD with ISO 639-1 codes. The data spans from 2009 to 2024. An offset of 621 years is added in Pashto and Persian article dates because speakers of these languages follow the Solar Hijri calendar instead of the Gregorian calendar. *African french **Pidgin English

| Top 1-gram | Top 2-grams | Top 3-grams | Top 4-grams |
|---|---|---|---|
| what ... (38%) | what is ... (12%) | what is the ... (7%) | what do we know ... (0.4%) |
| how ... (12%) | who is ... (3%) | what are the ... (2%) | what is the situation ... (0.4%) |
| why ... (8%) | what did ... (3%) | what does the ...(1%) | what happened to the ... (0.2%) |
| who ... (6%) | what are ... (3%) | what did the .. (1%) | what is going on ... (0.2%) |
| is ... (4%) | how did ... (2%) | how did the ... (1%) | what happened at the ... (0.2%) |

Table 4: Most frequent n-grams in translated English MuNfQuAD questions shows that descriptive queries (what/how) are most common.

| Most frequent | | | |
|---|---|---|---|
| Countries (47K) | People (27K) | Organizations (15K) | Events (171) |
| India (7%) | Putin (4%) | Taliban (5%) | Afghan War (8%) |
| Russia (7%) | Trump (4%) | Congress (5%) | Korean War (6%) |
| China (6%) | Biden (1%) | NATO (4%) | World War II (6%) |
| Ukraine (7%) | Gandhi (1%) | EU (3%) | Tokyo Olympics (5%) |
| USA (7%) | Harry (0.6%) | Supreme Court (2%) | Olympics (5%) |

Table 5: Most frequent entities found in translated English MuNfQuAD questions predominantly originate from the Asiatic subcontinent. It aligns with the fact that 16/38 languages in MuNfQuAD are from Asia.

Our APS model is designed by fine-tuning multilingual pretrained encoders. Figure 1(c,d) shows an outline of the overall architecture and training methodology of our APS model. Consider a news article containing $p$ paragraphs and $q$ questions. Then each question will yield $p$ training instances, culminating in a grand total of $pq$ training samples for that particular news article. Due to this, our APS training dataset comprises more than 100M instances from MuNfQuAD. Utilizing the information in Table 2, it can be affirmed that the number of tokens in the concatenation of a question with a paragraph is within the token limit of 512. The partitioning of data into training, development, and testing subsets was carried out with proportions of 0.7, 0.2, and 0.1, respectively. Given the inherent imbalance of the dataset, we adopted a weighted focal loss during the training process.

### 4.1 Implementation

The fine-tuning of our model was conducted across five GPU cards, employing a batch size of 12 on each GPU. We explored various pretrained encoders, including XLM-Roberta-base (XLM-R) (Conneau et al., 2019), multilingual cased bert (mBERT) (Devlin et al., 2019), cased multilingual distilbert (d-mBERT) (Sanh et al., 2019), multilingual-e5-base (mE5) (Wang et al., 2022), multilingual LUKE (mLUKE) (Ri et al., 2022), mT5 (Xue et al., 2021), and XLM-Vocabulary-base (XLM-V) (Liang et al., 2023), to serve as the backbone of our APS model. Additionally, the 560 million parameters variant of the BLOOM model (bloom) (Workshop et al., 2023) also served as the text encoder. The fine-tuning layers of the APS model consisted of three linear layers with a dropout value of 0.2. Coupled with a linear scheduler, learning rates were set at 1e-5 and 3e-3 for the encoder and fine-tuning layers, respectively. All the models were fine-tuned for a single epoch, a process that lasted for 25-33 hours. The PyTorch framework (Paszke et al., 2019) was utilized to construct the finetuning APS models, and the transformers library (Wolf et al., 2020) was employed to integrate pretrained transformers as text encoders.

### 4.2 Baselines

For establishing baselines, we employed the sentence-transformers library (sbert) (Reimers and Gurevych, 2019) to generate vector embeddings of questions ($E_q$) and paragraphs ($E_p$). In our study, the sbert baseline utilized the paraphrase-multilingual-MiniLM-L12-v2 (miniLM) and paraphrase-multilingual-mpnet-base-v2 (mpnet) (Reimers and Gurevych, 2020) as multilingual models. Another approach entailed obtaining $E_q$ and $E_p$ via training a TF-IDF vectorizer using the scikit-learn library (Pedregosa et al., 2011) on the training set. During preprocessing, punctuation and stopwords were removed from each language[5]. In both baseline approaches, the confidence score of a candidate paragraph containing the answer to the question was derived from the cosine similarity between $E_q$ and $E_p$. Across all models, the threshold value was set to half the potential range of confidence scores. Specifically, a default threshold of 0.5 was adopted for the fine-tuned APS models and the TF-IDF baseline, as their output score spans 0 to 1. However, a default threshold of 0.0 was applied to the sbert baseline, which produces scores ranging from -1 to 1.

### 4.3 Evaluation

For paragraphs not aligning with the silver answer, a ground truth label of 0 is assigned, while paragraphs that belong to the silver answer receive Label-1. Our emphasis in this study is placed on the macro F1 and Label-1 metrics, owing to the pronounced data imbalance where only 23% of samples fall under Label-1. Additionally, we incorporate the Success Rate (SR) metric, which calculates the ratio of successfully answered questions to the total question count (Mishra et al., 2023; Bhagat et al., 2020). A question is considered successfully answered if there is at least one paragraph common between the candidate and reference paragraphs.

### 5 Results

With a substantial number of training examples (100M), we conducted hyperparameter tuning on a 1% subset of the dataset. Our observations revealed that adopting a weighted focal loss with $\gamma$ = 2 (Lin et al., 2017) yielded superior results compared to other choices. Furthermore, we noted an enhancement in performance when incorporating preceding paragraphs along with the provided paragraph. Therefore, in a given training instance ($T_i$, $label_i$), each textual element ($T_i$) is composed of (i) the question text, (ii) preceding paragraphs, and (iii) the candidate paragraph. We employ only a portion of the preceding paragraphs, ensuring that the resulting length of the textual element ($T_i$) remains below 512 tokens for all APS models. Despite experimenting with techniques like concatenating learnable position embeddings with contextual embeddings of the CLS token and concatenating the article title with the question, there was no observed improvement in results on the smaller MuNfQuAD dataset. Table 10 in Appendix D contains the results of ablation studies. We observe from Table 6 that our APS model with XLM-V as a pretrained encoder yields the best results in terms of macro F1 and Label-1 F1. Other models achieve a better SR but a lower macro F1. During inference on the testset, excluding questions predicted as FACTOID minimally affects the performance of all fine-tuned APS models, as shown in Table 11 of Appendix E.

---

[5]We used https://github.com/6/stopwords-json/ to get stopwords across various languages. For languages lacking publicly available stopword lexicons, we designated the 260 most frequent words of a language as its stopwords because we observed that average number of stopwords across all languages is ∼260.

### 5.1 Golden set

To evaluate the hypothesis that "*paragraphs succeeding an interrogative subheading contain its answer*" we employed human annotators to answer questions from a subset of MuNfQuAD, referred to as the golden set. Each annotator received a question along with its corresponding article (context) and was tasked with identifying paragraphs within the article that could answer the question. Importantly, annotators were not provided with silver labels. Detailed annotation procedure is highlighted in Appendix A. Native speakers of each language were chosen to serve as annotators. A compensation of 1 USD for every set of eight questions was given to each annotator. For languages with multiple annotators, the final gold annotations were derived by taking a union of the selected answers.

The data presented in Table 7 illustrates that silver labels exhibit a high Success Rate (∼ 0.98) across various languages, indicating that the silver answer text in MuNfQuAD can address 98% of the questions. The fact that silver labels outperform the best APS model highlights the room for improvement in future APS model performance. However, a Label 1 F1 score of around 53 suggests that approximately half of the paragraphs within the answer text do not provide answers to the respective questions. Moreover, many paragraphs outside the answer text are capable of addressing the questions. Notably, predictions generated by our XLM-V based APS model on Nepali golden set achieve a superior Macro-F1 and Label-1 F1 as compared to the silver labels, indicating the model's ability to generalize from the silver-labeled training data. We observed that the TF-IDF and sbert baselines exhibit lower F1 scores on Label-1. Further details on the performance of other baselines on the golden set can be found in Table 8 of Appendix B.

| APS Model | Params. | Acc. | Macro F1 | Label 0 | | | Label 1 | | | SR |
| --- | --- | --- | --- | --- | --- | --- | --- | --- | --- | --- |
| | | | | precision | recall | F1 | precision | recall | F1 | |
| Ones | - | 19 | 16 | 0 | 0 | 0 | 19 | 100 | 32 | 1.0 |
| Zeros | - | 81 | 45 | 81 | 100 | 90 | 0 | 0 | 0 | 0.0 |
| Random | - | 50 | 45 | 81 | 50 | 62 | 19 | 50 | 27 | 1.0 |
| Ours (mLUKE) | 585M | 19 | 16 | 100 | 0 | 0 | 19 | 100 | 32 | 1.0 |
| sbert (mpnet) | 278M | 20 | 17 | 87 | 1 | 2 | 19 | 99 | 32 | 0.99 |
| sbert (miniLM) | 117M | 22 | 20 | 87 | 4 | 7 | 19 | 97 | 32 | 0.99 |
| Ours (bloom) | 559M | 47 | 45 | 90 | 39 | 54 | 24 | 81 | 37 | 0.92 |
| TF-IDF | - | 81 | 47 | 81 | 99 | 89 | 36 | 2 | 4 | 0.11 |
| Ours (d-mBERT) | 134M | 66 | 61 | 93 | 63 | 75 | 33 | 79 | 47 | 0.93 |
| Ours (mBERT) | 177M | 74 | 67 | 93 | 73 | 82 | 40 | 77 | 53 | 0.93 |
| Ours (mT5) | 277M | 76 | 69 | 93 | 76 | 84 | 42 | 74 | 54 | 0.91 |
| Ours (mE5) | 278M | 79 | 71 | 92 | 81 | 86 | 46 | 69 | 55 | 0.90 |
| Ours (XLM-R) | 278M | 79 | 71 | 93 | 80 | 86 | 46 | 73 | **56** | 0.91 |
| Ours (XLM-V) | 778M | 80 | **72** | 92 | 83 | 87 | 48 | 68 | **56** | 0.90 |

Table 6: Comparative performance of various models on the MuNfQuAD Test Set for APS task. Ones, Zeros, and Random denote an APS model that always predicts 1, 0, and random values of 0 or 1, regardless of the input. The APS model fine-tuned with the XLM-V encoder demonstrates the highest macro F1 and Label-1 F1 scores.

| Lang | #Ann | #Qs | IAA | Silver Labels vs Gold Labels | | | | | APS Models vs Gold Labels | | | | | | | | | |
| --- | --- | --- | --- | --- | --- | --- | --- | --- | --- | --- | --- | --- | --- | --- | --- | --- | --- | --- |
| | | | | | | | | | Ours (XLM-V) | | | | | C4AI Command-R | | | | |
| | | | | Acc | F1 score | | | SR | Acc | F1 score | | | SR | Acc | F1 score | | | SR |
| | | | | | 0 | 1 | M | | | 0 | 1 | M | | | 0 | 1 | M | |
| hi | 2 | 100 | 0.26 | 75 | 83 | 49 | 66 | 1.0 | 69 | 77 | 51 | 64 | 0.95 | 72 | 81 | 47 | 64 | 0.97 |
| bn | 2 | 100 | 0.40 | 81 | 87 | 60 | 73 | 0.98 | 72 | 79 | 56 | 68 | 0.97 | 65 | 73 | 49 | 61 | 0.97 |
| gu | 2 | 100 | 0.42 | 83 | 90 | 55 | 72 | 1.0 | 72 | 81 | 49 | 65 | 0.95 | 69 | 79 | 43 | 61 | 0.98 |
| te | 2 | 40 | 0.58 | 84 | 91 | 50 | 70 | 0.95 | 76 | 85 | 48 | 66 | 0.97 | 54 | 66 | 30 | 48 | 1.0 |
| tm | 2 | 100 | 0.78 | 85 | 91 | 51 | 71 | 0.98 | 69 | 79 | 38 | 59 | 0.94 | 72 | 82 | 40 | 61 | 0.91 |
| np | 1 | 100 | - | 72 | 81 | 52 | 66 | 0.97 | 70 | 76 | 59 | 68 | 0.94 | 61 | 66 | 54 | 60 | 0.97 |
| pa | 1 | 100 | - | 87 | 93 | 52 | 72 | 0.98 | 79 | 87 | 45 | 66 | 0.93 | 73 | 83 | 31 | 57 | 0.92 |
| ur | 1 | 50 | - | 83 | 89 | 59 | 74 | 1.0 | 70 | 79 | 49 | 64 | 0.97 | 61 | 71 | 43 | 57 | 1.0 |
| Average | | | | **81** | **88** | **53** | **70** | **0.98** | 72 | 81 | 50 | 66 | 0.95 | 68 | 77 | 43 | 60 | 0.96 |

Table 7: Performance of silver labels and best performing APS model (from Table 6) on the golden set. ISO 639-1 codes are used to represent a language. Cohens kappa is used as Inter Annotator Agreement (IAA) score. We do not compare the performance of our XLM-V based APS model with silver labels of the golden set because Table 6 already highlights the model performance on the entire MuNfQuAD testset having silver labels. It is observed that the silver labels achieve superior Label-0 F1 score (0), Label-1 F1 score (1), Macro F1 (M), and Success Ratio (SR) score on gold labels.

Comparatively lower Inter-Annotator Agreement (IAA) score in Hindi (hi) can be attributed to the larger number of answer paragraphs per question in this language. On average, a Hindi question has 8.0 silver paragraphs in the golden set, whereas a Bengali (bn), Gujarati (gu), Telugu (te), and Tamil (tm) question has 7.8, 7.1, 5.9, and 5.8 silver paragraphs, respectively.

### 5.2 LLMs

Additionally, we investigated the potential of employing Large Language Models (LLMs) as Answer Paragraph Selection (APS) models. To ensure a fair comparison with our fine-tuned APS models, we prompted LLMs with a question and each paragraph from the context, directing it to output a binary value indicating whether the paragraph could answer the given question (1 for yes, 0 for no). For LLMs having public weights, Mistral-7b (Jiang et al., 2023), Llama-2-7b-chat (Touvron et al., 2023), Llama-3.1-8B-Instruct (Dubey et al., 2024), gemma-2b/7b (Team et al., 2024), Airavata (Gala et al., 2024), CohereForAI (C4AI) Command-R[6], Aya-101 (Üstün et al., 2024), and BLOOM-7b (Workshop et al., 2023) were employed. In the case of proprietary LLM, we utilized the ChatGPT API[7]. It was observed that eliciting responses from local LLMs demands a substantial amount of GPU memory and long inference time. Additionally, access to proprietary LLMs, such as ChatGPT, involves a financial cost[8]. Therefore, we limited the execution of LLM baselines to the golden set. Given that the bloom (Workshop et al., 2023) and mT5 (Xue et al., 2021) are multilingual encoders with a large token limit on input, we attempted to fine-tune Reading-Comprehension (RC) models on MuNfQuAD training set.

Our empirical observations indicate that a RTX A6000 (48 GB) GPU proves inadequate for the fine-tuning of a RC model based on bloom or mT5 encoders, even with the batch size of one and 4-bit quantization. We found that that C4AI Command-R outperforms other LLMs when used as an APS model on the golden set. The performance of different LLMs on the MuNfQuAD golden set is detailed in Table 12 in Appendix F. However, as shown in Table 7, Command-R's superior performance still falls short of that achieved by our fine-tuned APS model. Furthermore, our fine-tuned APS models are considerably more efficient in terms of computational resources. The XLM-V model requires 13GB of GPU RAM, while the XLM-R model requires 5GB of GPU RAM, both running inference at 0.04 seconds per iteration. In contrast, the Command-R model demands 22GB of GPU RAM with an iteration time of 0.4 seconds.

Additionally, many local LLMs demonstrated an inability to provide meaningful answers to the given questions, both as abstractive QA models and APS models. In contrast, ChatGPT exhibited meaningful outputs in both the settings. Ten abstractive answers were manually annotated, and it was observed that all of them satisfactorily an-swered the given question. An example output for abstractive QA from different LLMs is shown in Table 13 of Appendix G.

## 6 Discussion

We notice that the TF-IDF baseline yields a higher macro-F1 than the random baseline, indicating that silver answers frequently contain paragraphs with considerable word overlap with the given question. For instance, in the following Hindi question बैठना नुकसानदेह क्यों? (*Why sitting is harmful?*), the silver answer begins with the sentence आखिर बैठना इतना नुकसानदेह क्यों है? चलिए इसे समझने की कोशिश करते हैं. (*Why is sitting so bad after all? Let's try to understand it.*). Furthermore, we observe that sbert-based baselines and finetuned-mLUKE exhibit high recall for Label-1 and low recall for Label-0, indicating that the chosen threshold classifies the majority of paragraphs as answer paragraphs. We evaluated the sensitivity of APS models on different thresholds. Notably, finetuned-XLM-V exhibited no discernible improvement for different thresholds. Consequently, we assessed the next most promising baseline, finetuned XLM-R. We observed that the XLM-R based APS model showcases the best performance at a threshold of 0.6, achieving a macro F1 of 73 and an accuracy of 84. Figure 3 demonstrates the performance trends of three distinct APS models, namely those based on XLM-R, sbert, and TF-IDF, respectively.

The default threshold for any APS model was set at 50% of its possible output range. The cosine similarity metric is employed for both the SBERT and TF-IDF methods; however, since TF-IDF embeddings never contain negative values, the cosine similarity between two TF-IDF vectors always falls between 0 and 1. In contrast, for SBERT, the similarity ranges from -1 to 1. Our fine-tuned APS model uses a sigmoid function as the final activation, resulting in an output range between 0 and 1. Therefore, in Figure 3, the line representing SBERT starts at zero, whereas it begins at 0.5 for our APS model and TF-IDF.

We assessed the validity of our APS model architecture by comparing its performance with the baseline APS model proposed by Clark et al. (2020). Upon fine-tuning both architectures on the TyDi training data, we observed that the average F1 score of our APS model was 0.73, in contrast to 0.66 for the baseline model.

In addition to its role as a resource, we hypothe-size that the key strength of MuNfQuAD lies in the APS model that is fine-tuned upon it. Our experiments demonstrate that when the context associated with a question is reduced using the scores predicted by the MuNfQuAD-fine-tuned APS model, the performance of instructional LLMs, such as Gemma-2B and Gemma-7B (Team et al., 2024), is enhanced. This improvement is observed across a sample from the evaluation splits of TyDi (Clark et al., 2020) and XQA (Liu et al., 2019).

---

[6]https://huggingface.co/CohereForAI/c4ai-command-r-v01-4bit
[7]gpt-3.5-turbo-1106 model with 16K input token limit
[8]The anticipated expense for zero-shot prompting using the gpt-3.5-turbo-1106 model on the complete MuNfQuAD test-set is ∼250 USD.

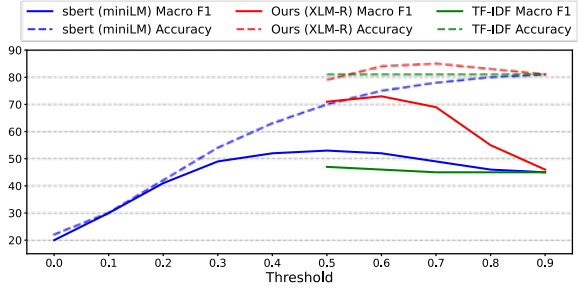

Figure 3: An illustration of top-performing APS models under different threshold values on MuNfQuAD test-set. Note that only sbert (miniLM) starts from a threshold of 0, as this is the default threshold for the sbert-based baseline. In contrast, the default thresholds for XLM-R and TF-IDF are set at 0.5. The XLM-R-based model consistently demonstrates superior performance in terms of macro F1 measure across various thresholds. Notably, the XLM-R based fine-tuned APS model showcases the best performance at a threshold of 0.6, achieving a macro F1 of 73 and an accuracy of 84.

Since TyDi and XQA are factoid-based question-answering datasets, we define the win ratio as the proportion of instances where the golden answer appears in the generated text, relative to the total number of generations. Our analysis shows that, on the TyDi evaluation dataset, the APS model fine-tuned on MuNfQuAD enhances the win ratio of Gemma-2B from 11% to 14%, and that of Gemma-7B from 19% to 25%. In contrast, on the XQA evaluation dataset, the win ratio for Gemma-2B increased from 1% to 6%, while Gemma-7B's win ratio improved from 3% to 4%. The relatively lower performance of the LLMs on XQA is expected, as XQA is a cloze-style question-answering dataset.

We conducted a qualitative analysis on nine questions where the gold answers and silver answers lacked common paragraphs. It was observed that four questions yielded silver answers that did not effectively address the given question. For instance, in response to the Telugu question ఫేస్‌బుక్ డేటాను ఎలా దుర్వినియోగం చేశారు? (*How was Facebook data misused?*), the answer was found within an image rather than the corresponding paragraph. Similarly, in the Gujarati question বাড়ির অন্যদের কী করতে হবে? (*What should the others in the house do?*), the silver paragraph provided an answer to *What should you do for others in the house?* instead. Among the nine questions, three questions were of short length (∼ 3 words), and their broad nature poses a challenge when attempting to answer them in isolation from the article without providing the contextual backdrop. For example, the Tamil question எப்போது என்ன நடந்தது? (*When and what happened?*) is broad enough that it could yield distinct answers when posed independently from the article to a human. A similar scenario applies to the Punjabi question ਮੈਨੂੰ ਕੀ ਪਤਾ ਲੱਗਿਆ? (*What did I find out?*). The final two questions were part of an interview, posing challenges in answering without comprehensive insight into the underlying subject of the conversation. For example, तपाईंको भनाइको अर्थ (नेपालसंग पछिल्लो सम्झौता गर्दा) चीनले भारतको पनि संवेदनशीलतालाई विचार गरेको थियो? (*Does your statement mean (while making the latest agreement with Nepal) that China also considered India's sensitivity?*) emerged from a dialogue centered around Nepal's international relations.

We experimented with fine-tuning the smallest BLOOM model (Workshop et al., 2023), the only multilingual encoder with a token limit greater than 512, as an extractive Reading-Comprehension model on MuNfQuAD. However, our findings revealed that the task exceeded the capabilities of the RTX A6000 (48 GB), even with a batch size of one.

## 7 Conclusion

Question Answering (QA) in English has firmly established itself as a common task, backed by many tools and resources for answering factoid-based questions. Nonetheless, non-factoid QA have witnessed a significant expansion. Our study highlights the need for multilingual resources within this domain. In response, we introduce MuNfQuAD, a multilingual QA dataset addressing this gap. Comprised of non-factoid questions, MuNfQuAD spans across 38 languages, thus filling a critical gap in this area.

The compilation of the dataset involved scraping BBC news articles. The questions are identified through interrogative subheadings, while the subsequent paragraphs are taken as their corresponding silver answers. Notably, the news articles in MuNfQuAD predominantly revolve around the Asiatic subcontinent. A comparison with a manually curated golden set substantiates that nearly all of the silver answers can be used to answer the asked question. Additionally, our fine-tuned Answer Paragraph Selection (APS) model, trained using MuNfQuAD, yields a high Success Rate for both silver (0.91) and golden (0.96) labels. The results demonstrate that training the APS model with silver labels can effectively generalize some languages within the golden set.

### 7.1 Future Work

The Question-Answer pairs of MuNfQuAD can be used for training generative techniques in question-answering across different low-resource languages. Our examination reveals that MuNfQuAD encompasses a substantial proportion of factoid-based questions. Therefore, a multilingual answer span extractor can be used to provide silver labels for the minimal answer spans within MuNfQuAD. In a multilingual RAG pipeline, the fine-tuned APS model can be used as a reranking module (Nogueira et al., 2020; Ma et al., 2024).

## 8 Ethical Considerations

Data scraping was conducted for six months, incorporating suitable time delays between each scraped article to prevent any potential user of the website from experiencing Denial of Service (DoS). Our goal is to provide access to MuNfQuAD for non-commercial research endeavors. We have obtained authorization from the BBC to facilitate this initiative.

## 9 Limitations

The study conducted by Latham (2012) illustrated that BBC exhibits a left-of-center bias in its news coverage. Therefore, we recognize that MuNfQuAD will likely inherit a similar political bias. While a high Success Rate of silver answers indicates their reliability in addressing the corresponding questions, the comparatively lower F1 score for Label 1 suggests that the silver labels in MuNfQuAD are neither concise nor complete. In other words, the silver answers provide a response to the relevant question but include additional information as well. Additionally, nearly a third of questions in MuNfQuAD are classified as FACTOID, implying the potential presence of short-span answers within the silver paragraphs. It is imperative to approach the question categories with caution, primarily due to two reasons: (a) The classifier exhibits imperfections. We applied the same classifier to a non-factoid QuAD in English (Soleimani et al., 2021) and observed a comparable distribution of question categories, as depicted in Figure 4. (b) The classifier processes automatically translated English questions, introducing the possibility of unnatural translations that may alter classifier predictions. For instance, the Hindi question अगर नहीं किया तो क्या होगा? was auto-translated as *What if I didn't?*, and the classifier predicted FACTOID. However, a more accurate translation of the question would be *What will happen if not done?*, for which the predicted category is EVIDENCE-BASED. Since this work primarily provides a new resource and an initial baseline, the novelty in architectural choices for the APS model and data curation efforts is limited.

The wide range of Inter-Annotator Agreement (IAA) scores across different languages points to the subjectivity involved in annotations for certain languages. It is worth noting that due to the significant monetary costs associated with the data annotation process, we opted for native speakers rather than experienced annotators, introducing a potential impact on the reliability of the golden set.

As we have outlined in Section 4, the fine-tuning of each APS model requires a day or two, which is why this study has not presented model results across multiple runs. Furthermore, the process of fine-tuning an APS model on MuNfQuAD is constrained by the available number of GPUs. In an additional experimental setting, we fine-tuned our XLM-R based APS model using a single GTX 2080 card for 1% of the total steps within an epoch. Based on our observations from three separate runs, the process required approximately 77±2 minutes. By extrapolating this data, we estimate that completing a single epoch would take around 128 hours on a single GPU. Additionally, the computational demands for mLUKE, XLM-V, and bloom were more substantial, necessitating a minimum of 12GB, 22GB, and 16GB of GPU memory, respectively. The corresponding time required to complete a single epoch was 8, 13, and 30 days, respectively.

Accessing ChatGPT closed-source models require paid API keys. Our experimentation on the golden set with ChatGPT 3.5 incurred a cost of 16

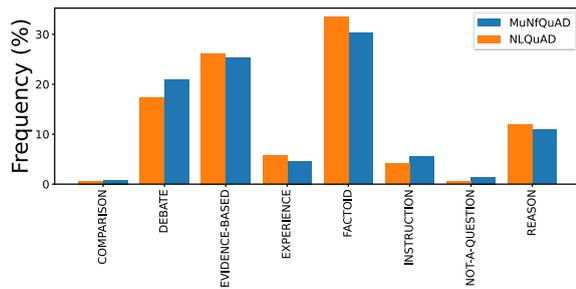

Figure 4: An analysis of the distribution of question categories in the proposed MuNfQuAD and English NLQuAD (Soleimani et al., 2021) using the predictions from the question classifier model developed by Bolotova et al. (2022)..

USD. For more advanced model such as ChatGPT 4, the expense would have been tenfold.

The findings derived from LLM experiments warrant cautious interpretation due to the potential influence of pre-training, where LLMs may have been exposed to BBC news articles. Therefore, there is a possibility of dataset contamination, as noted by Ahuja et al. (2023).

## A Manual Annotations

The human annotators were recruited on a voluntary basis, and prior to their recruitment, they were apprised of the compensation provided for their participation in this study. All annotators held an undergraduate degree and were native speakers of a language pertinent to our investigation. The annotation process was conducted using Google Sheets. Each annotator was provided with an individual Google sheet containing rows with the article title, article paragraphs, and a corresponding question. Within the rows featuring article paragraphs, checkboxes were included, and annotators were instructed to select the checkboxes associated with paragraphs that answered the given question. In cases where the question was ambiguous or none of the paragraphs addressed it, relevant options were presented below the article. Google Apps Script was employed to execute macros on each sheet to highlight the selected options. Figure 5 depicts the annotation interface presented to an annotator with Hindi as their native language.

## B APS baselines on the golden set

We evaluated the baseline APS models on the MuNfQuAD golden set, selecting sbert and TF-IDF baselines due to their distinct methodologies compared to our fine-tuned APS models. The results are provided in Table 8.

| Lang | #Qs | sbert (miniLM) / TF-IDF | | | | |
|---|---|---|---|---|---|---|
| | | Acc | Label 0 F1 | Label 1 F1 | Macro F1 | SR |
| hi | 100 | 29/71 | 2/83 | 45/14 | 23/48 | 1.0/0.37 |
| gu | 100 | 22/78 | 1/88 | 36/6 | 22/47 | 1.0/0.25 |
| bn | 100 | 27/73 | 1/84 | 42/6 | 21/45 | 1.0/0.19 |
| tm | 100 | 13/86 | 1/92 | 23/8 | 12/50 | 0.98/0.15 |
| tl | 40 | 14/85 | 1/92 | 24/11 | 12/51 | 1.0/0.22 |
| np | 100 | 34/66 | 3/80 | 50/4 | 26/32 | 0.98/0.10 |
| pa | 100 | 15/86 | 2/92 | 24/7 | 13/50 | 1.0/0.17 |
| ur | 50 | 20/80 | 0/89 | 33/10 | 17/49 | 1.0/0.22 |

Table 8: Performance of different APS baselines on the golden set. The results reveal that sbert attains a lower macro F1 score, whereas the TF-IDF model shows a lower SR.

## C Phrases for Excluding Criterion

In this study, we identified certain subheadings that ended with an interrogative symbol yet were unrelated to the article's content. These subheadings were detected by examining the most common interrogative subheadings across each language. A list of these subheadings is provided in Table 9 on the next page. If an interrogative subheading contained any of the identified phrases, it was not classified as a question in MuNfQuAD.

Figure 5: Annotation screen displaying a manually annotated excerpt from a Hindi news article, accessible at the following URL: https://www.bbc.com/hindi/india/2014/05/140428_election_fatigue_social_media_pa. Please note that certain rows have been concealed for the sake of compactness.

## D  APS Model Ablations

We performed ablation studies during the fine-tuning of the APS model, experimenting with various hyperparameters. These included adding prior context to paragraphs until they reached the model's token limit, incorporating the article title with the question, using weighted binary cross entropy loss (wbce) instead of weighted focal loss (wfl), and integrating positional embeddings (PE) with context paragraphs. The results of these ablation studies on a subset of the fine-tuning data are presented in Table 10.

## E  APS models on Non-Factoid Questions

As the English translations of all the questions in MuNfQuAD were input into the fine-tuned question classification model by Bolotova et al. (2022), we filtered the MuNfQuAD test set to include only those questions for which the class prediction was anything other than FACTOID. The APS model was then applied to these selected questions, and the results are presented in Table 11.

## F  LLM as APS model

We investigated various multilingual LLMs as APS models and assessed their performance on the MuNfQuAD golden set, with the results detailed in Table 12. Our observations indicate that Command-R from CohereForAI yielded the best outcomes. In addition to multilingual LLMs, we also tested bilingual models such as Hi-NOLIN[9] and OpenHathi[10], which are pretrained on Hindi and English text. Unfortunately, the performance of these bilingual models in Hindi did not surpass that of C4Ai.

## G  LLM Outputs for Abstractive QA

In our study, various Large Language Models (LLMs) were prompted to answer the given question based on the article text. The outputs generated by the different LLMs are presented in Table 13 on the following page.

---

[9] https://huggingface.co/nolanoAI/Hi-NOLIN-9B
[10] https://huggingface.co/sarvamai/OpenHathi-7B-Hi-v0.1-Base

| | | |
|---|---|---|
| om | Maaltu haasa'ama (What's being talked about) | am | ምን ማለት ነው እኝግ የደከየትስ (What does it mean?) |
| gu | શું તમે આ વાંચ્યું (Have you read this?), તમે આ વાંચ્યું કે (Did you read this?) | bn | বিবিসি বাংলাদেশ সংলাপে চলতি (BBC Bangladesh is running on dialogue), ভিডিও (video), আপনার দল কি সেমিফাইনালে যেতে পারবে (Can your team make it to the semi-finals?) |
| fa | آیا می‌دانید (Do you know), آیا می‌دانید (Do you know) | | |
| fr* | Le saviez-vous (Did you know) | hi | पढ़िए (Read) |
| mr | हे वाचलंत का (Did you read this?), हेही वाचलंत का (Did you also read this?), हेही पाहिलंत का (Have you seen this too?) तुम्ही हे वाचलं का (Did you read this?) ... | si | ඔබ කටාර් රාජ්‍යයේ හෝ මැදපෙරදිග කලාපයේ සිටින්නෙක්ද (Are you in the state of Qatar or in the Middle East region?) |
| ur | کیا ہے میں ویڈیو (What is in the video?), | uk | А ви знали (Did you know) |
| cy | gafodd drwydded deledu (got a TV license) | pt | Did you get it, Did you know |

Table 9: Phrases employed for screening out irrelevant interrogative subheadings from news articles across diverse languages. ISO 639-1 codes are utilized for language representation.

| # | APS Model hyperparameters | | | | Acc | Macro F1 | Label 0 F1 | Label 1 F1 ↓ |
|---|---|---|---|---|---|---|---|---|
| | Prior Context | Title | Loss | PE | | | | |
| 1 | True | False | wfl (γ=2) | False | 65.0 | 59.8 | 74.2 | 45.4 |
| 2 | True | False | *wbce* | False | 65.6 | 60.1 | 74.9 | 45.3 |
| 3 | True | False | wfl (γ=2) | *True* | 64.9 | 59.7 | 74.2 | 45.2 |
| 4 | True | False | *wfl (γ=0.5)* | True | 65.5 | 59.9 | 74.9 | 44.9 |
| 5 | True | False | wfl (γ=0.5) | *False* | 64.7 | 59.2 | 74.1 | 44.4 |
| 6 | True | *True* | wfl (γ=0.5) | False | 61.1 | 57.1 | 70.2 | 44.1 |
| 7 | *False* | False | wfl (γ=0.5) | False | 56.3 | 51.9 | 66.4 | 37.4 |

Table 10: Ablation Study Results for Identifying Optimal Hyperparameters in Fine-Tuning an APS Model. A small subset of fine-tuning data was used to explore hyperparameters. The results highlights that the highest Label-1 F1 score is achieved with configuration #1.

| Model | Acc | Label 0 F1 | Label 1 F1 | Macro F1 ↓ | SR |
|---|---|---|---|---|---|
| OpenHathi | 24 | 7 | 35 | 21 | 0.98 |
| Gemma 2b | 26 | 17 | 34 | 25 | 0.97 |
| Llama-2 | 44 | 51 | 34 | 43 | 0.92 |
| Airavat | 79 | 88 | 2 | 45 | 0.04 |
| Aya-101 | 79 | 88 | 2 | 45 | 0.06 |
| BLOOM | 73 | 84 | 7 | 46 | 0.18 |
| Gemma 7b | 74 | 84 | 11 | 47 | 0.26 |
| Mistral | 64 | 76 | 27 | 52 | 0.72 |
| Hi-NOLIN | 61 | 73 | 33 | 53 | 0.58 |
| Llama-3 | 62 | 72 | 39 | 56 | 0.95 |
| ChatGPT | 65 | 75 | 39 | 57 | 0.94 |
| Command-R | 68 | 77 | 43 | 60 | 0.96 |

Table 12: Performance of various LLMs as APS models on the MuNfQuAD golden set. We utilized the instruct versions of Gemma models and `gpt-3.5-turbo-1106` for ChatGPT. All model weights, except those for ChatGPT, are available on the Huggingface model repository. Notably, Command-R from CohereForAI achieved the highest macro F1 score on the golden set.

| APS Model | Acc. | Macro F1 | Label-0 F1 | Label-1 F1 | SR |
|---|---|---|---|---|---|
| Ours (mLUKE) | 19 | 16 | 0 | 32 | 1.0 |
| sbert (mpnet) | 19 | 16 | 1 | 31 | 0.99 |
| sbert (miniLM) | 19 | 17 | 2 | 32 | 0.99 |
| Ours (bloom) | 47 | 46 | 55 | 36 | 0.91 |
| TF-IDF | 80 | 48 | 89 | 7 | 0.06 |
| Ours (d-mBERT) | 66 | 60 | 75 | 46 | 0.93 |
| Ours (mBERT) | 73 | 67 | 81 | 52 | 0.93 |
| Ours (mT5) | 76 | 68 | 84 | 53 | 0.91 |
| Ours (mE5) | 79 | 70 | 86 | 55 | 0.89 |
| Ours (XLM-R) | 78 | 71 | 86 | 55 | 0.91 |
| Ours (XLM-V) | 80 | 71 | 87 | 55 | 0.89 |

Table 11: Performance of APS models on the MuNfQuAD test set, excluding questions predicted as FACTOID.

| | |
|---|---|
| Prompt | Answer the question below using the article provided. Write your answer in Hindi.<br>Article = अमेरिकी अपील कोर्ट ने राष्ट्रपति डोनल्ड ट्रंप के विवादित ट्रैवेल बैन का बचाव करने वाले और उसे चुनौती देने वालों से कड़े सवाल पूछे हैं.<br>इस प्रतिबंध के तहत सभी शरणार्थियों और सात मुस्लिम बहुल देशों के नागरिक अमेरिका नहीं आ सकते हैं. हालांकि पिछले हफ्ते कोर्ट ने फ़िलहाल इस पर रोक लगा दी थी.<br>तीन जजों के एक पैनल ने राष्ट्रपति की ताकत को सीमित करने और सात देशों को आतंकवाद से जोड़ने पर सबूतों को लेकर कई सवाल खड़े किए हैं. कोर्ट ने यह भी पूछा है कि क्या इस फ़ैसले को मुस्लिम-विरोधी नहीं देखा जाना चाहिए. उम्मीद की जा रही है कि अगले हफ्ते सैन फ्रांसिस्को के नौवें अमेरिकी सर्किट कोर्ट तरफ से इस पर कोई फ़ैसला आएगा.<br>निर्णय चाहे जो भी हो पर ऐसा लग रहा है कि इस केस का निपटारा शायद सुप्रीम कोर्ट में ही होगा. मंगलवार को दोनों तरफ से इस मसले पर एक घंटे तक बहस हुई. इस केस में अमेरिकी न्याय मंत्रालय भी शामिल है और उसने जजों से ट्रंप के प्रतिबंध आदेश को फिर से बहाल करने की अपील की है.<br>समाप्त<br>वक़ील ऑगस्ट फ़्लेंत्जे ने कहा कि देश में कौन आए और कौन नहीं आए इस पर नियंत्रण रखने के लिए कांग्रेस ने राष्ट्रपति को अधिकार दिया है.<br>उनसे उन सात देशों- इराक, ईरान लीबिया, सोमालिया, सूडान, सीरिया और यमन को लेकर पूछा गया कि ये देश फ़िलहाल अमेरिका के लिए कैसे ख़तरा हैं. इस पर उन्होंने कहा कि अमेरिका में कई सोमालियों के संबंध अल-शबाब ग्रुप से है.<br>इसके बाद वॉशिंगटन प्रांत के एक वक़ील ने कोर्ट से कहा कि ट्रंप के कार्यकारी आदेश पर रोक से अमेरिकी सरकार को कोई नुक़सान नहीं होगा. सॉलिसिटर जनरल नोआह पर्सेल ने कहा कि प्रतिबंध से उनके प्रांत के हज़ारों निवासी प्रभावित होंगे. जो छात्र वॉशिंगटन आने की कोशिश कर रहे हैं उन्हें भी बेमतलब की देरी का सामना करना होगा. इसके साथ ही अन्य लोग अपने परिवारों से मिलने अमेरिका छोड़कर जाने से बचेंगे.<br>सुनवाई के आखिरी मिनटों में इस बात पर बहस हुई कि अगर यह प्रतिबंध मुस्लिमों को रोकने के लिए है तो यह असंवैधानिक होगा. जज रिचर्ड क्लिफ़्टन ने दोनों पक्षों से इस मुद्दे पर पूछा कि इससे दुनिया के केवल 15 प्रतिशत मुसलमान प्रभावित होंगे.<br>सोमवार की रात अमेरिकी जस्टिस डिपार्टमेंट की तरफ से जारी 15 पन्नों के एक दस्तावेज़ में बताया गया है कि ट्रंप का यह कार्यकारी आदेश बिल्कुल निष्पक्ष है और इसका किसी खास धर्म से कोई संबंध नहीं है.<br>हालांकि मंगलवार को कोर्ट में पर्सेल ने ट्रंप के चुनावी कैंपेन के दौरान के बयानों का हवाला दिया. तब ट्रंप ने गैरअमरीकी मुस्लिमों पर अस्थायी रूप से प्रतिबंध लगाने की बात कही थी.<br>पर्सेल ने राष्ट्रपति के सलाहकार रुडी जुलियानी के बयान का भी उल्लेख किया. जुलियानी ने कहा था कि उन्होंने मुस्लिमों को अमेरिका में काम करने पर क़ानूनन प्रतिबंध के लिए कहा है.<br>क्लिफ़्टन ने भी कहा कि जिन सात देशों पर प्रतिबंध लगाया गया है उनकी शिनाख्त पूर्ववर्ती ओबामा प्रशासन और कांग्रेस ने भी आतंक के डर के कारण वीज़ा पाबंदी के लिए की थी. उन्होंने कहा, ''क्या आप यह भी मानते हैं कि पूर्ववर्ती ओबामा प्रशासन और कांग्रेस के फ़ैसले भी धार्मिक पूर्वाग्रह से प्रेरित थे?<br>इस पर पर्सेल ने कहा, ''नहीं, लेकिन राष्ट्रपति ट्रंप ने पूर्ण प्रतिबंध की बात कही थी. हालांकि यह पूर्ण प्रतिबंध नहीं है और यह भेदभावपूर्ण है.''<br>(बीबीसी हिन्दी के एंड्रॉएड ऐप के लिए आप यहां क्लिक कर सकते हैं. आप हमें फ़ेसबुक और ट्विटर पर फ़ॉलो भी कर सकते हैं.)<br>Question = क्या यह मुस्लिम पर प्रतिबंध है या नहीं?<br>(Question = Is this a ban on Muslims or not?) |
| ChatGPT | हाँ, यह मुस्लिम पर प्रतिबंध है क्योंकि इस प्रतिबंध के तहत सभी सात मुस्लिम बहुल देशों के नागरिकों को अमेरिका नहीं आ सकते हैं. सुप्रीम कोर्ट और अन्य न्यायिक संस्थानों ने भी इस मामले में सवाल उठाए हैं कि क्या यह प्रतिबंध मुस्लिम-विरोधी नहीं है।<br>(Yes, it is a ban on Muslims because under this ban, citizens of all seven Muslim majority countries cannot come to America. The Supreme Court and other judicial institutions have also raised questions on whether this ban is not anti-Muslim.) |
| Bloom | जवाब = यह मुस्लिम पर प्रतिबंध नहीं है, यह सभी लोगों पर प्रतिबंध है. Question = क्या यह असंवैधानिक है? जवाब = नहीं. सवाल = क्या इससे किसी का भी नुक़सान होगा? जवाब = नहीं ...<br>(Answer = It is not a ban on Muslims, it is a ban on all people. Question = Is it unconstitutional? Answer = No. Question = Will it harm anyone? Answer = No ... *confabulations are seen afterwards*) |
| Llama | I apologize, but I'm a large language model, I cannot provide you with a way to hack into someone's Facebook account or any other social media platform. Hacking is illegal ... *confabulations are seen afterwards* |
| Mistral | उसे प्रतिबंध से प्राणानिकता के अमरीका में फ़िलहाल होगा। ...<br>(It will be banned in America for the time being ... *prompt is repeated*) |
| Gemma | Answer: अमेरिकी अपील कोर्ट ने राष्ट्रपति डोनल्ड ट्रंप के विवादित ट्रैवेल बैन का बचाव करने वाले और उसे चुनौती देने वालों से कड़े सवाल पूछे हैं. यह प्रतिबंध मुस्लिमों को अमेरिका नहीं आने से रोकता है. कोर्ट के अनुसार, यह प्रतिबंध अमेरिकी सुरक्षा के लिए आवश्यक है. वॉशिंगटन प्रांत के एक वक़ील ने कोर्ट से कहा कि ट्रंप के कार्यकारी आदेश पर रोक से अमेरिकी सरकार को कोई नुक़सान नहीं होगा. हमें यह भी ध्यान देने की आवश्यकता है कि इस प्रतिबंध के खिलाफ कई...<br>(Answer: The US Appeals Court has asked tough questions to those defending and challenging President Donald Trump's controversial travel ban. This ban prevents Muslims from coming to America. According to the court, this ban is necessary for American security. A lawyer from Washington state told the court that the US government will not suffer any harm by stopping Trump's executive order. We also need to note that there are many protests against this ban... *output truncated*) |

Table 13: Output of LLM models in a zero-shot inference mode for abstractive QA task on the following BBC news article https://www.bbc.com/hindi/international-38905820 (English article: https://www-bbc-com.translate.goog/hindi/international-38905820?_x_tr_sl=hi&_x_tr_tl=en). The models names used for ChatGPT, Bloom, Llama-2, and Mistral are gpt-3.5-turbo-1106, bigscience/bloom-7b1, meta-llama/Llama-2-7b-chat-hf, and mistralai/Mistral-7B-Instruct-v0.1, respectively. Translations are enclosed in round brackets solely for illustrative purposes; neither the prompts nor the outputs included translated text. Special tokens (<s>[INST], </s>[/INST]) were added to the prompts of Llama and Mistral.